\documentclass[letterpaper, 10 pt, conference]{ieeeconf}  

\IEEEoverridecommandlockouts                              

\usepackage{graphicx} 
\usepackage{tabularx}
\usepackage[noadjust]{cite}
\usepackage[caption=false,font=footnotesize]{subfig}
\usepackage{amsmath} 
\usepackage{amssymb}  
\usepackage{adjustbox}
\usepackage{wrapfig}
\usepackage{float}
\usepackage{hyperref}
\usepackage{orcidlink}
\usepackage{booktabs}
\DeclareMathOperator*{\argmax}{arg\,max}

\title{Disentangling perception and reasoning for improving data efficiency in learning cloth manipulation without demonstrations}

\author{Donatien~Delehelle\orcidlink{0009-0000-0049-6585}, Fei~Chen\orcidlink{0000-0003-4397-0931}, Darwin~Caldwell\orcidlink{0000-0002-6233-9961}%
\thanks{Darwin Caldwell is with Advanced Robotics, Istituto Italiano di Tecnologia (IIT), 16163 Genoa, Italy (e-mail: {\tt\small darwin.caldwell@iit.it})}
\thanks{Fei Chen is with epartment of Mechanical and Automation Engineering, T-Stone Robotics Institute, The Chinese University of Hong Kong, Hong Kong (e-mail: {\tt\small f.chen@ieee.org})}
\thanks{Donatien Delehelle is with the University of Genova, 16145, Italy, and Advanced Robotics, Istituto Italiano di Tecnologia (IIT), 16163 Genoa, Italy (e-mail: {\tt\small donatien.delehelle@iit.it})}
}

\begin{document}

\maketitle
\thispagestyle{empty}
\pagestyle{empty}

\begin{abstract}

Cloth manipulation is an ubiquitous task in everyday life, but it is still an open challenge for robotics. The difficulties in developing cloth manipulation policies are attributed to the high-dimensional state space, complex dynamics, and high propensity to self-occlusion displayed by fabrics. As analytical methods haven't been able to provide robust and general manipulation policies, reinforcement learning (RL) is seen as a promising approach to these problems. However, in order to address the large state space and the complex dynamics, data-based methods usually rely on big models and long training times. The inferred computational cost significantly hampers the development and adoption of these methods.
Additionally, because of the challenge of robust state estimation, garment manipulation policies usually adopt an end-to-end learning approach with workspace images as input.~While this approach allows for a conceptually trivial sim-to-real transfer via real-world fine-tuning, it also impairs a significant computational cost on the training on agents with a very lossy representation of the environment state.
This paper aims at questioning this common design choice by exploring an efficient and modular approach to RL for cloth manipulation. We show that through careful design choices, we can significantly reduce model size and training time when learning in simulation. Additionally, we showcase how our optimal simulation-based model can be transferred to the real-word. We evaluate our work on the Softgym benchmark and achieve a significant performance improvements over available baselines on our task, with a significantly smaller model.

\end{abstract}


\section{INTRODUCTION}
\PARstart{R}{obotic} manipulation of deformable objects, particularly garments, remains a crucial and largely unsolved challenge with broad societal relevance, especially in assistive and medical contexts. As populations age, the demand for robotic support in activities of daily living (ADLs) continues to increase. Although manipulation of rigid and articulated objects has progressed rapidly, deformable-object manipulation lags behind due to the high-dimensional configuration space, severe self-occlusion, and nonlinear dynamics.

Deep neural networks have shown strong capabilities in high-dimensional settings, and recent progress in RL has enabled more effective learning through interaction. This has yielded promising results in cloth manipulation~\cite{matas2018sim, wu2019learning, jangir2020dynamic, lin2021softgym, chen2024trakdis, salhotra2022learning}, yet RL’s sample inefficiency remains a major obstacle. Cloth simulation is slower than rigid-body physics, and the action space itself scales with the number of material points. Consequently, unlike locomotion tasks, where high-dimensional observations coexist with low-dimensional actions, the combinatorial sampling complexity of state-action pairs for picking is exponentially higher than that imposed by the configuration-space dimension alone.

Several works mitigate this issue by leveraging expert demonstrations~\cite{seita2020deep, matas2018sim, jangir2020dynamic, chen2024trakdis}, effectively constraining exploration to neighbourhoods of successful trajectories. While efficient, these approaches offer limited guarantees regarding scalability to new tasks or manipulation scenarios.

A further limitation is that nearly all prior work relies on end-to-end image-based policies mapping workspace images to 2D pick-and-place actions. Although this facilitates sim-to-real fine-tuning, it entangles perception, exploration, and reasoning, making it difficult to attribute performance gains to specific components of the system.

In this paper, we instead approach cloth manipulation from an exploration-centred perspective. To address the challenges posed by the large configuration-space dimensionality, we disentangle perception and reasoning by training a manipulation agent directly from full state information in simulation.

Our framework is built on four principles.~\textbf{Offline pre-training}: heuristic rollouts efficiently populate diverse regions of the configuration space.~\textbf{Multi-objective training}: structured objectives promote robust latent representations~\cite{kumar2023offline}.~\textbf{Full-state exploitation}: complete geometric information induces strong inductive biases, enabling rapid identification of key invariants such as the leverage of corner grasps.~\textbf{Q-level sim-to-real transfer}: instead of transferring visual encoders, we distil simulation-trained Q-functions into real-world vision-based policies via supervised training, enabling reuse of the same simulation agent across multiple setups.

We evaluate our approach on the cloth-spreading task, which spans a wide range of cloth configurations. Our contributions are threefold: (i) we demonstrate substantial performance gains with a compact model trained directly in configuration space and without demonstrations; (ii) we analyse in detail the impact of each design principle; and (iii) we introduce a cross-modality distillation strategy enabling practical sim-to-real transfer.
\noindent Additionally, code and supplementary material will be shared on the \href{https://ddonatien.github.io/mopt-website/}{project's website}.

\section{RELATED WORKS}
\subsection{Data-driven methods for cloth manipulation}
Learning-based methods have shown strong performance in cloth manipulation tasks by leveraging demonstrations, simulation, and vision-based policies. Early work by~\cite{PIGNAT201761} demonstrated a data-efficient approach for assistive dressing using Gaussian mixture models fit to demonstrations in task-relevant frames.~\cite{seita2020deep} used simulation to train vision-based imitation policies for cloth flattening, while~\cite{wu2019learning, hoque2020visuospatial} employed goal-conditioned policies and visuospatial dynamics models for pick-and-place flattening using RGB-D inputs. These model-free approaches enable flexible manipulation directly from perception.~\cite{zhang2022learning} further extend this paradigm by combining vision models and motion primitives to dress a bed-ridden manikin with an hospital gown initially hanged from a pole.
Other methods incorporate explicit structure to improve generalization.~\cite{weng2022fabricflownet} leverage optical flow prediction to guide folding, while~\cite{lin2022learning} learn a mesh-based dynamics model over visible cloth points.~\cite{yan2021learning} adopt contrastive objectives for latent dynamics modeling to improve cloth flattening. In high-speed settings,~\cite{ha2022flingbot} combine vision-based grasp point selection with dynamic motion primitives for cloth unfolding. Across these approaches, simulation, structured representations, and policy learning all play a central role in managing cloth’s high-dimensional, nonlinear behavior. In this work, we build on this foundation by directly leveraging full-state simulation data to pretrain compact and transferable Q-function representations.

\subsection{Offline Reinforcement Learning and Pretraining}
Offline reinforcement learning (RL) trains agents from fixed datasets without environment interaction~\cite{levine2020offline}. While appealing for safety and cost, offline RL faces distributional shift issues, which are mitigated through conservative objectives~\cite{kumar2020conservative} or implicit regularization~\cite{nair2021awacacceleratingonlinereinforcement}.
Combining offline pretraining with online fine-tuning has shown strong performance in robotics. For example,~\cite{kalashnikov2018scalable} use large-scale offline Q-learning followed by online adaptation to achieve robust grasping. This offline-to-online pipeline has been shown to improve sample efficiency~\cite{ball2023efficientonlinereinforcementlearning, xie2022policyfinetuningbridgingsampleefficient}.
Inspired by the generalization capabilities of foundation models, recent work explores leveraging pretrained representations for downstream control~\cite{yang2023foundationmodelsdecisionmaking}.~\cite{kumar2023offline} show that large-capacity offline RL models scale well and generalize across tasks.
We adopt this pretraining paradigm to learn a compact state representation from cloth simulation data, which accelerates policy learning and improves generalization to new tasks.

\section{BACKGROUND}
\begin{figure*}[t]
  \centering
  \includegraphics[width=.85\linewidth]{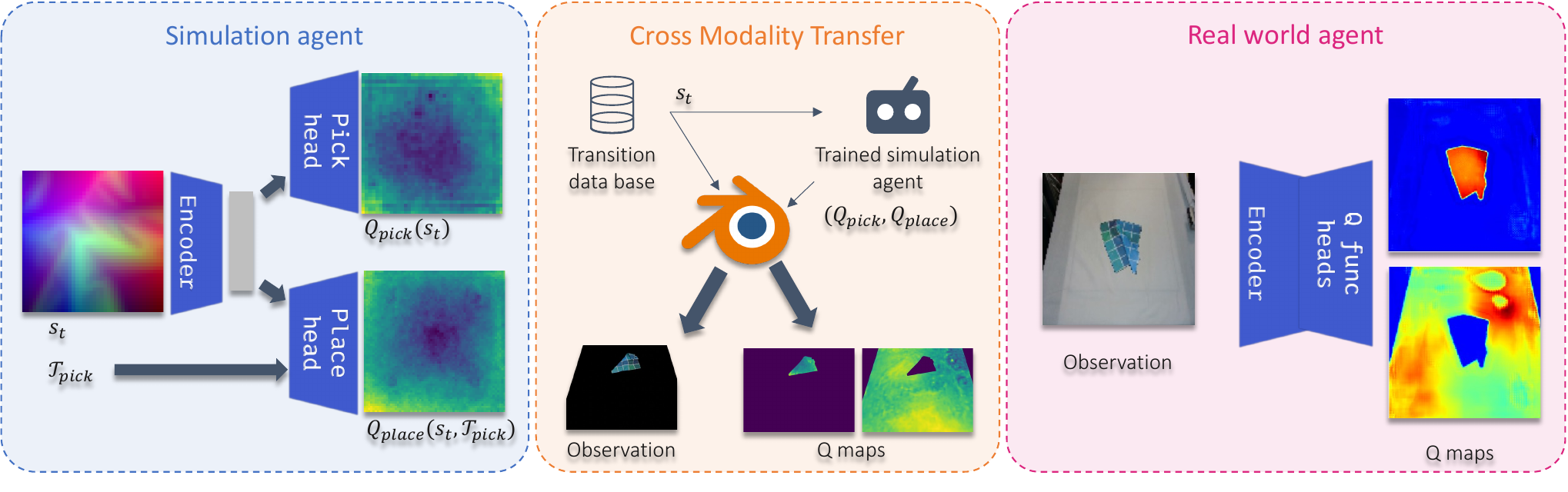}
  \caption{Overview of our method. We train an optimal agent in simulation with a maximal access to the cloth's state. When transfering our agent to the real world, we distill the knowledge from the state-based simulation agent to a vision-based real world agent through the ad-hoc generation of a dataset, densely labelled by the simulation agent}%
  \label{fig:meth_img}
\end{figure*}

We begin by outlining the fundamentals of value-based reinforcement learning, with a focus on the Double DQN algorithm.

\subsection{Value-Based Reinforcement Learning and Double DQN}

We model the cloth manipulation task as a Markov Decision Process (MDP), defined by the tuple $(\mathcal{S}, \mathcal{A}, \mathcal{P}, r, \gamma)$, where $\mathcal{S}$ is the state space, $\mathcal{A}$ the action space, $\mathcal{P}$ the transition dynamics, $r: \mathcal{S} \rightarrow \mathbb{R}$ the reward function, and $\gamma \in (0, 1)$ the discount factor. A policy $\pi: \mathcal{S} \rightarrow \mathcal{A}$ maps states to actions, and the agent aims to find the optimal policy $\pi^*$ that maximizes the expected discounted return from any initial state $s_0$.

The return, denoted $G_\pi(s_0)$, is the total accumulated reward when following policy $\pi$ starting from $s_0$:

\begin{equation}
\label{eq:optpol}
\begin{split}
\pi^* &= \argmax_{\pi} G_\pi\\
\text{where} \quad G_\pi : s_0 &\mapsto \sum_{t=0}^\infty \gamma^t r(s_{t+1} \mid a_t = \pi(s_t))
\end{split}
\end{equation}

Q-learning optimises the action-value function $Q^\pi(s, a)$, which estimates the expected return when taking action $a$ in state $s$ and thereafter following policy $\pi$:

\begin{equation}
\label{eq:qfunc_pb}
\begin{split}
Q^\pi(s, a) &= \mathbb{E}[G_\pi(s_0) \mid s_0 = s, a_0 = a]\\
\pi(s) &= \argmax_{a \in \mathcal{A}} Q^\pi(s, a)
\end{split}
\end{equation}

Although computing the infinite sum in Eq (\ref{eq:optpol}) is intractable, $Q^\pi$ can be approximated recursively using the Bellman equation:

\begin{equation}
\label{eq:bellman}
Q(s_t, a_t) = r(s_{t+1}) + \gamma \max_{a \in \mathcal{A}} Q(s_{t+1}, a)
\end{equation}

In Deep Q-Networks (DQN), the Q-function is approximated by a neural network. However, the use of the max operator in Eq (\ref{eq:bellman}) leads to overestimated Q-values. Double DQN~\cite{van2016deep, hasselt2010double} addresses this by decoupling action selection and evaluation: the online network $Q_\theta$ selects the action, while a target network $Q_{\theta'}$ evaluates it:

\begin{equation}
\label{eq:ddqn}
\begin{split}
Q_\theta(s_t, a_t) &= r(s_{t+1}) + \gamma Q_{\text{max}}(s_{t+1})\\
Q_{\text{max}}(s_{t+1}) &= \max_{a \in \mathcal{A}} Q_{\theta'}(s_{t+1}, a)
\end{split}
\end{equation}

Traditionnaly, the target and online network were periodically \textit{switched} during training. In practice, Polyak averaging on the network's weights is more commonly used.
\[\theta' \leftarrow \tau \theta + (1 - \tau) \theta'\]
With $\tau \in (0, 1)$ controling the update rate.

\section{METHOD}
The goal of the cloth flattening task is to maximise the area of the workspace covered by a piece of square cloth. The task starts from a random crumpled configuration, and the agent has to remove all folds and ripples through its manipulation actions.
The task is to be solved with the one-handed pick and place actions $a_t = (\mathcal{T}_{pick}, \mathcal{T}_{place})$. Similarly to~\cite{wu2019learning, zeng2020transporter}, we model the intrinsic dependencies of the pick and place actions by decomposing the problem into picking and a pick-conditioned placing:
\begin{equation}
\label{eq:pick_place_factor}
\pi: s \mapsto (\pi_{pick}(s), \pi_{place}(s, \mathcal{T}_{pick}))\\
\end{equation}

This results in learning the two Q functions expressed in Eq (\ref{eq:pick_place_qfunc}).

\begin{equation}
    \label{eq:pick_place_qfunc}
    \begin{split}
        Q_{pick}: s_t, a_t \mapsto &\max_{\mathcal{T}_{place}} Q_{place}(s_{t+1}, (\mathcal{T}_{pick}, \mathcal{T}_{place}))\\
        Q_{place}: s_t, a_t \mapsto &r(s_{t+1}) + \gamma \max_{\mathcal{T}_{pick}} Q_{pick}(s_{t+1}, \mathcal{T}_{pick})
    \end{split}
\end{equation}
In this section, we present our learning method for the efficient training of the cloth manipulation agent.
An overview of our method is shown in Fig.~\ref{fig:meth_img}.
The training is subdivided into three stages: offline pre-training, fine-tuning in simulation, and transfer to the real world.

\subsection{Optimal simulation agent}
\subsubsection{Training objective}%
\label{subsec:objectives}
The reward of the cloth flattening task is the value of the area covered by the cloth on the 2D ground plane.
Because this reward is a very lossy representation of the state, training the agent against this only objective presents a risk of mode collapse in the latent vector of the Q-network. This is the case because two different actions on the same state, or two inital states upon which the same action is applied can produce two new states with identical coverage.
Recovery from these collapsed states is possible if disambiguating samples are present in the training data. However, because of the high dimensionality of both the state and action spaces, encountering such transitions may be unlikely.
To mitigate this problem, we introduce auxiliary objectives during the pre-training stage:\\
\textbf{Cloth Fold Straight:} The goal of the cloth fold task is to arrange the cloth into a one-fold rectangle close to the centre of the scene. The reward is made up of two terms: one is the node-to-node Euclidean distance between the left and right sides of the cloth, and the second term is the Euclidean distance between the right side's nodes and their desired position close to the scene's centre.\\
\textbf{Cloth Fold Diagonal:} The goal of the cloth fold task is to arrange the cloth into a one-fold triangle close to the centre of the scene. The negative reward is made of two terms: one is the node-to-node Euclidean distance between the top-left side and the bottom-right sides of the cloth, and the second term is the Euclidean distance between the bottom-right side's nodes and their desired position close to the scene's centre.\\
We linearly transform all reward functions so that they are bounded in the $[0, 50]$ interval and that all tasks are solved by maximising the reward.

\subsubsection{State space}

\begin{figure}[ht]
  \centerline{\includegraphics[width=0.9\linewidth]{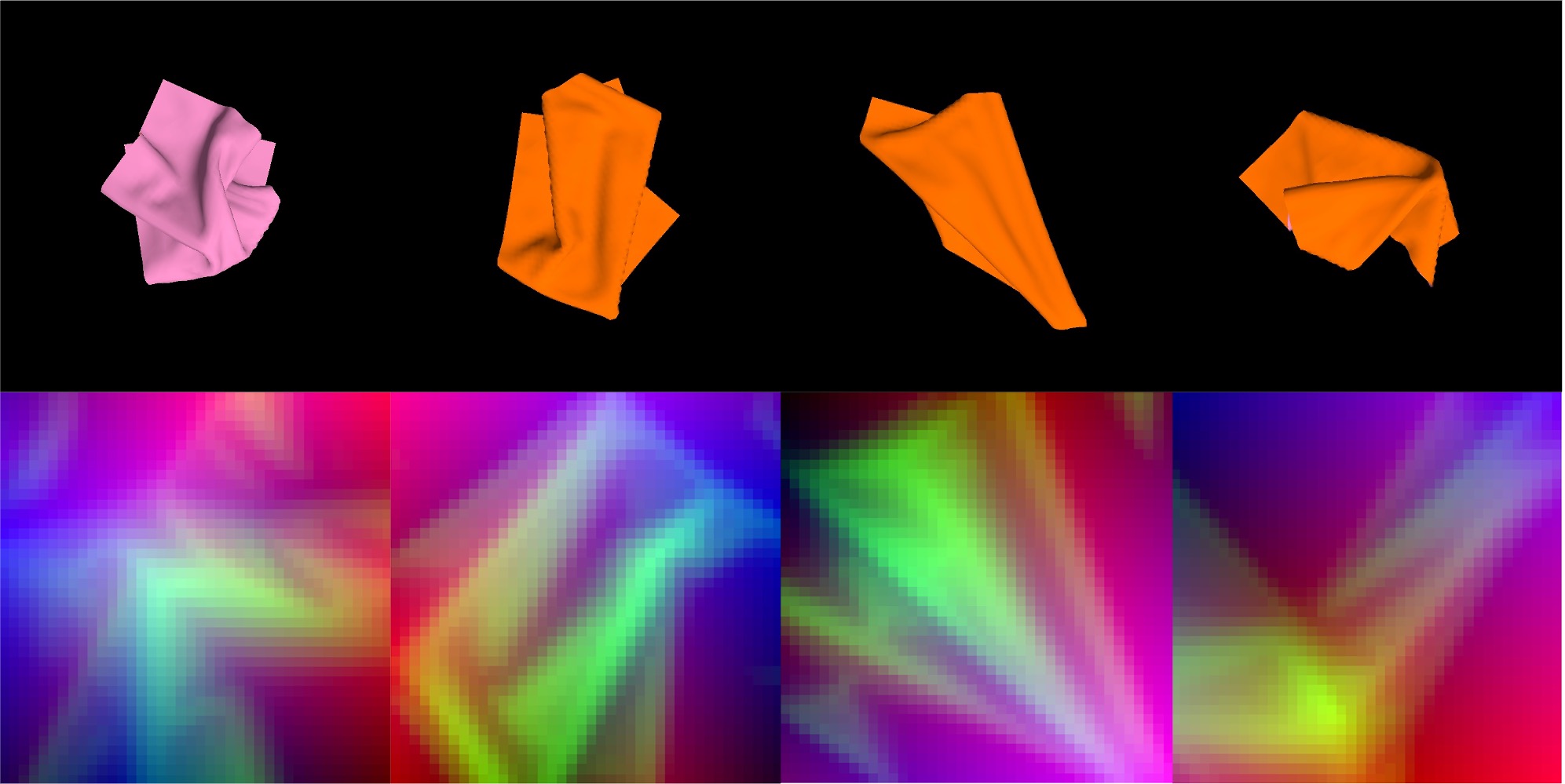}}
  \caption{Top row: Random states of the cloth. Bottom row: Corresponding \emph{state images}. The color of each pixel in the state image is computed from the Euclidean position of the corresponding particle in the cloth's particle grid. In order to maximise saturation in this figure, the state images presented are normalized along each channel.}\label{fig:st_img}
\end{figure}

We use the full configuration of the cloth (i.e.~the positions of the nodes of the cloth) as the MDP's state. As a result, we use interchangably the terms \textit{configuration} and \textit{state} even if these terms usually refer to different objects beyond the context of this paper. As all other elements in the scene, barring the effectors, are constant, this is the maximally informative state of the environment. As the dimension of such state is high (a vector of 4800 floats in our case), state-based RL approaches~\cite{wu2019learning, matas2018sim, clegg2020learning} traditionally sub-sample the state. Instead, we chose to represent the cloth's state as an image in order to leverage the efficient convolutional layers in our agent's model. We argue that, due to the local continuity of the positional information (i.e.~close nodes have similar positions in space) and the fixed grid structure, this \textit{state image}, with each of the 3 colour channels representing the positions of the nodes along the 3 axis of the scene's reference frame, is an advantageous representation as it conveys, by design, the intrinsic (i.e.~the relation of the nodes between each other) and the extrinsic (i.e.~the position of the nodes relative to the scene) positional information in a joint manner. We provide sample \textit{observation}/\textit{state image} pairs in Fig.~\ref{fig:st_img}. As convolution layers are equivariant with position, we pad the state images with on black pixel to provide a strong feature along the borders of the cloth.

\subsubsection{Action space}

Traditionally, pick-and-place actions are defined as pixel coordinates that are back-projected into 3D, which facilitates sim-to-real transfer since such operations can be executed directly from RGB-D observations. As our work focuses solely on simulation, we instead represent the picking action as a node index on the 2D grid defining the cloth discretisation.~This introduces a useful inductive bias, enabling the policy to identify high-leverage nodes such as corners, and prevents sampling actions outside the cloth, which has shown to accelerate early training. A resulting issue is that some nodes may be unreachable due to vertical stacking of cloth layers. To avoid storing invalid transitions, we record the actual node grasped by the simulator rather than the infeasible proposal. The placing action is represented separately as a 2D point on the normalised ground plane.

\subsubsection{Training}\label{subsec:offpt}
The training of our simulation agent is done in two stages. First, the agent is pre-trained on an offline dataset, then it is fine-tuned in simulation.
During the offline pre-training, the agent minimizes the sum of three objective functions as shown in Eq (\ref{eq:loss_fns}): pick and place losses (L2-loss on the objectives described in~\ref{subsec:objectives}) and a bounding loss. Indeed, as noted in~\cite{kumar2020conservative}, off-policy methods risk overestimating values for out-of-distribution states. This can lead to instability in training, particularly as our pick and place Q-values are interdependent. We mitigate this by bounding Q-values below the theoretical return maximum with the bounding loss shown in Eq (\ref{eq:loss_fns}). While simple, this solution proved effective in practice.

\begin{equation}
\label{eq:loss_fns}
\begin{split}a
\mathcal{L} =& \mathcal{L}_{pick} + \mathcal{L}_{place} + \mathcal{L}_{bound}\\
\mathcal{L}_{bound} = \quad & ||\max\left(\frac{R_{max}}{1-\gamma}, \max_a Q_{pick}(s_{t+1}, a_t)\right)||_2 \\
+ & ||\max\left(\frac{R_{max}}{1-\gamma}, \max_a Q_{place}(s_t, a_t)\right)||_2
\end{split}
\end{equation}

The online fine-tuning is done following an $\epsilon$-greedy exploration strategy with an un-prioritized replay buffer, pre-filled with transitions from the offline dataset.
More precisely, we use the tuple $(\epsilon_{pick}, \epsilon_{place})$ to describe the probability at which the corresponding agent action is replaced by a random action.
During the online fine-tuning, the bounding loss is dropped as the agent has the possibility to explore the high-quality actions and self-correct. Additionally, the pick and place loss terms for accessory objetives are ignored.

\subsubsection{Implementation details}
The agent's Q-network follows an encoder-decoder architecture with a shared convolutional encoder and one head for each Q function.
A detailed description of the Q network architecture is given in the supplementary material.
The encoder is made up of two convolutional layers followed by a linear layer. Each head is comprised of one linear layer and a fully convolutional decoder. All modules use layer norm and the GELU activation function. The `pick' head's input is made up of the output of the encoder, while for the `place' head, the pick action is concatenated.
The agent is trained following the double DQN algorithm with a Polyak averaging with a smoothing factor of $5e^{-4}$ for the target network. The discount factor is chosen to be 0.9. Finally, a termination clause is set when the agent reaches $95\%$ of the maximum coverage area.
Training is done, both offline and online, with the AdamW~\cite{loshchilov2017decoupled} optimizer.
The online fine-tuning loop starts with the collection of 2000 transitions followed by 4 optimization iterations with a batch size of 8192. The fine-tuning stage is made-up of 20 iteration of this loop. The size of the replay buffer is set at $S_{rb} = 100000$.
The values of all hyperparameters for all training stages are summed-up in the supplementary material.
Regarding the offline data, the dataset contains 6.5 million $(s_t, a_t, s_{t+1}, r_{t+1})$ tuples, with 6\% from the fold-to-unfold strategy from~\cite{wu2023learning} and the rest from random interactions. We split the data 80/20 between training and validation.
Finally, the state images are normalised during training by subtraction the mean and division by the standard deviation of the offline dataset.

\subsection{Transfer to the real-world}
\begin{figure*}[t]
  \centering
  \includegraphics[width=\textwidth]{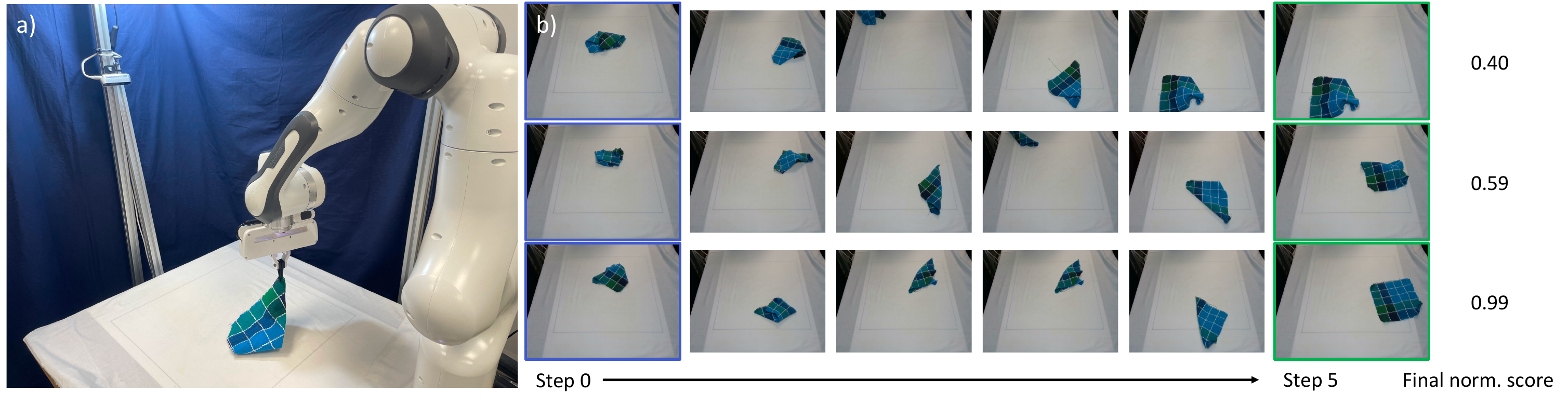}
  \caption{\textbf{(a)} Real-world setup for the cloth manipulation task.~\textbf{(b)} Successful examples of real-world roll-outs with corresponding normalised improvement}\label{fig:rw_things}
\end{figure*}
As the agent trained in simulation has access to the complete state of the cloth, and this state is not available in the real-world, it is no possible to directly transfer the learnt simulation agent to the real world. Instead, we train a new real-world agent through cross-modality knowledge distillation. This distillation process utilizes the optimal simulation agent as a source of ground truth for the real-world agent that is then trained by supervision.
This method relies on generating corresponding state-observations pairs of which the observation is used as an input to the real-world agent and the state as an input to the optimal simulation agent which, in turn generates the corresponding \textit{ground truth} Q-values.
The observation are generated by importing simulation states inside a Blender scene closely matching the real-world setup shown in Fig.~\ref{fig:rw_things} (a). The observations are generated from 100K states sampled from the offline dataset and states encountered during online training. The appearance of the scene and the cloth are manually configured in Blender via custom-generated assets. More details on the data generation process is given in the supplementary material.
As the teacher uses a different modality (i.e.~states) thant the student, we dub this knowledge distillation process as \textit{cross-modality}.
The real-world's agent architecture is a UNet with a two-channels head for, respectively, the pick quality values and the place quality values. The model is trained with the Adam~\cite{kingma2014adam} optimiser against a l2-loss with the labels provided by the simulation agent.
While the data-generation process can be time consuming on certain hardware, the supervised training is much faster than the RL stage. Indeed, the labels provided by the simulation agent are dense, allowing us to use a reconstruction loss which is more informative than the signal from the Bellman error.
Finally, we highlight that our real-world agent can adapt to workspace modification by only re-generating the supervision data and no re-training of the optimal simulation agent.

\section{EXPERIMENTAL RESULTS}

We evaluate our method in simulation on the \textit{Softgym}~\cite{lin2021softgym} environment, a popular benchmarking environment for cloth manipulation.
We compare our results with three baselines: MVP~\cite{wu2019learning}, VCD~\cite{lin2022learning} and Deformable affordance~\cite{wu2023learning}, which are the reference methods for the one-handed pick-and-place cloth flattening task on the softgym benchmark with freely available code at the time of writing.
Each agent is tested over 100 episodes. The initial state of the episode is randomly generated following the heuristic provided by the \textit{Softgym}~\cite{lin2021softgym} environment.\\
We report, as metrics, the mean performance over the episodes. Additionally, following recommandations from~\cite{agarwal2021deep}, we report the inter-quartile mean (IQM).
The performance of the agent is measured as the normalized improvement, which is the delta of scene coverage between the first and the list timestep divided by delta of the maximal possible improvement.
For a score of 1, the cloth in completely flattened from any initial configuration. If the cloth is more crumpled after the episode, the score is negative.

\subsection{Benchmarking results}
An overview of the results are given in Table~\ref{tab:res_tab}. As Deformable Affordance and MVP have an actor critic architecture, we only report the number of learnable parameters in the critic's network for a fair comparison. VCD relies on a connectivity prediction network and a dynamics prediction network for its planning. We thus report the sum of the parameter count of both networks in our table. We observe a 21\% improvement over the best baseline in IQM with a 95\% smaller network. This results demonstrates that despite the high dimensionality of the cloth's state space, the reasoning problem behind cloth flattening can be addressed with much smaller models than previously thought.
The marginal improvement from the fine-tuned model over the zero-shot performance of the offline-trained model hints toward a major influence of the size of the offline dataset on the performance of the agent. This hypothesis is explored in more depth in~\ref{sec:ablation}.

\begin{table*}[t]
  \centering
  \caption{Performances on the cloth flatten task.~\textbf{le} stands for \textbf{linear encoder}, \textbf{ce} for \textbf{convolutional encoder}, \textbf{zs} for \textbf{zero shot} and \textbf{ft} for \textbf{fine tuned}}\label{tab:res_tab}
  \begin{tabularx}{\textwidth}{l X X X | X X r}
  \toprule
  \textbf{Metric}&\textbf{Ours (le, zs)}&\textbf{Ours (ce, zs)}&\textbf{Ours (ce, ft)}&\textbf{MVP}&\textbf{VCD}&\textbf{Deformable Affordance}\\
  \midrule
    n params & 1.27M & 0.98M & 0.98M & 1.82M  & 5.47M & 18.87M\\
  \midrule
    IQM & 0.715 & 0.896 & \textbf{0.913} &~-~& 0.632 & 0.756\\
    mean & 0.803 & 0.805 & \textbf{0.846} & 0.3 & 0.603 & 0.674\\
  \bottomrule
  \end{tabularx}
\end{table*}

\subsection{Ablation studies}\label{sec:ablation}
\begin{figure}[ht]
  \centering
  \subfloat[]{
  \includegraphics[width=0.3\linewidth]{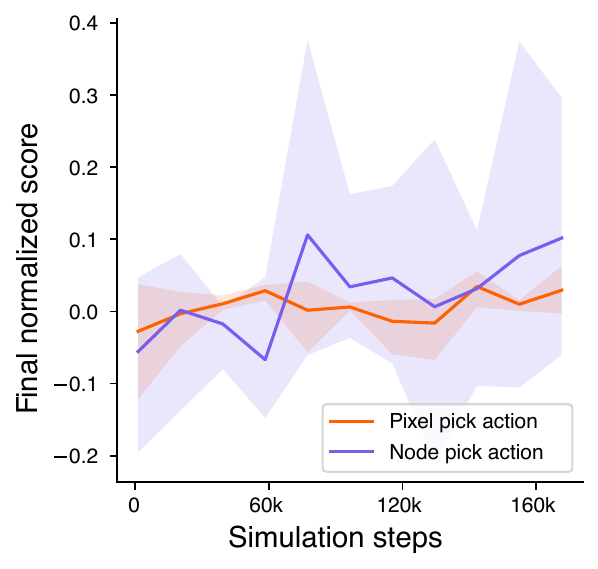}%
  \label{fig:pxvnd}}
  \hfil
  \subfloat[]{
  \includegraphics[width=0.3\linewidth]{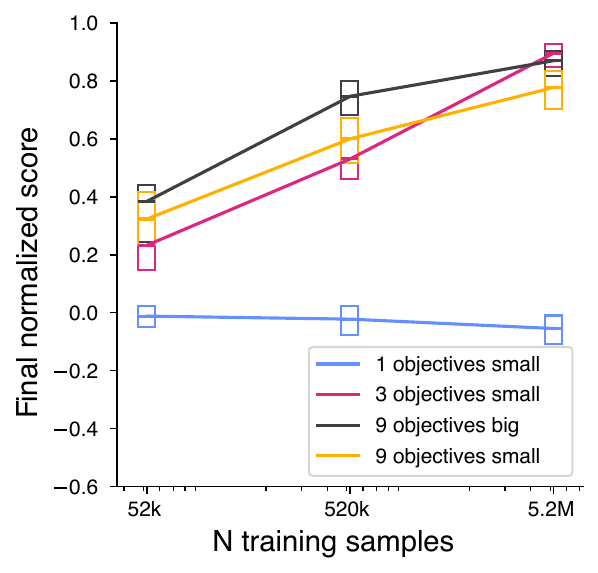}%
  \label{fig:datasize}}
  \hfil
  \subfloat[]{
  \includegraphics[width=0.3\linewidth]{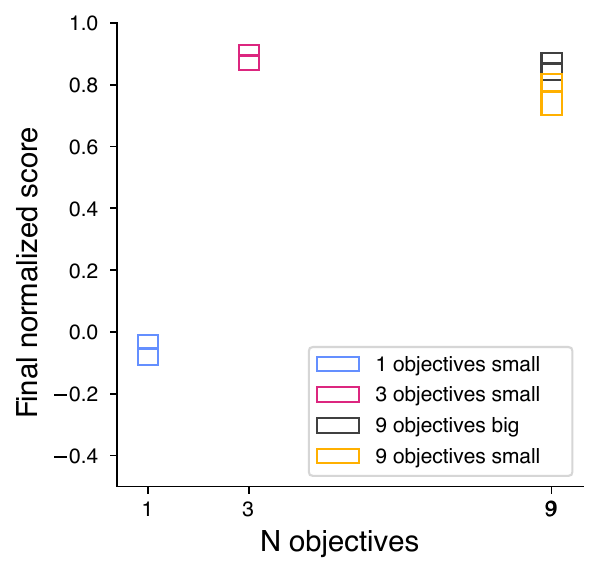}%
  \label{fig:nfunc}}
  \caption{Qualitative evaluations of the training parameters.~\textbf{(a)} Sample efficiency plot of pixel-space picking and node-space picking.~\textbf{(b)} Scaling law of data for offline pre-training.~\textbf{(c)} Impact on multi-objective training on final performance. For~\textbf{(b)} and~\textbf{(c)} the mean performance value is represented with its bootstrapping confidence interval}\label{fig:res_figs}
\end{figure}

\textbf{Does the node picking action improve sample efficiency?}\\
To assess the impact of picking mesh nodes instead of 2D positions, we compare two agents trained for 190,000 simulation steps, differing only in their picking strategy. In the pixel-picking variant, the agent may select points outside the cloth, incurring a -10 penalty. We average four training runs per pick modality; Fig.~\ref{fig:pxvnd} reports the mean return, with the envelope showing min–max values. Results indicate faster and more exploratory learning with node picking, leading to improved sample efficiency and broader return distributions.

\textbf{How does the multi-objective pre-training impacts performance?}\\
To assess the impact of multi-objective pre-training, we compare zero-shot performance after offline training using one, three, and nine objective functions during training. The additional objectives are derived from the auxiliary objectives detailed in Subsection~\ref{subsec:objectives} by folding along the other perpendicular axes. As shown in Fig.~\ref{fig:nfunc}, performance improves from one objective to three objectives but saturates beyond that. We hypothesized that this is due to saturation in the decoder or in the latent space. To test this, we trained a larger \textit{big} model with similar outcomes, suggesting diminishing returns from additional objectives above three.

We investigate the scaling laws of the dataset size by evaluating agents pre-trained on smaller, randomly sampled, sub-datasets of 1\%, 10\%, and 100\% of the original dataset's side. Results in Fig.~\ref{fig:datasize} reveal an logarithmic relationship between dataset size and zero-shot performance, with a stronger performance in high objective counts in low data setting, further supporting our earlier hypothesis.

\textbf{How does the convolutional encoder impact performance?}\\
We compare zero-shot performance between agents using an MLP encoder versus a convolutional encoder in Table~\ref{tab:res_tab}. We observe that the convolutional encoder slightly outperforms the linear one. While the performance gap is modest, the linear encoder remains 30\% larger in its minimal size, i.e~only two layers with first layer of the input's size and the second layer of the latent vector's size. This experiments shows that our convolutional architecture together with the \emph{state image} allows for a more compact, weight-efficient, network, without sacrificing performance.

\subsection{Real-world experiment}
We evaluate our method on a real-world setup composed of a tabletop Franka-Emika Panda arm with a custom, near-punctual gripper. The manipulated object is a 21cm × 21cm piece of cotton cloth placed within a 70cm × 70cm workspace, matching the scale used in simulation. A top-down Intel RealSense D435 camera, placed opposite the robot, captures RGB-D images used for both action selection and evaluation. The setup and some representative trajectories are shown in Fig.~\ref{fig:rw_things}.

To compute the normalised improvments, the camera image is rectified with a perspective transform, and the cloth mask is extracted, for the first and last timesteps, via HSV thresholding. The coverage score is defined as the ratio of white pixels in the cloth mask to that of a fully flattened cloth.

\begin{table}
\begin{center}
  \caption{\small Final normalised improvement in real-world experiment}\label{tab:rw_tab}
  \begin{tabularx}{\linewidth}{ l X r }
  \toprule
  \textbf{Policy}&\textbf{mean}&\textbf{std}\\
  \midrule
  Random                & 0.349 & 0.254 \\
  Deformable Affordance & \textbf{0.683} & -\\
  Ours~(offline)        & 0.523 & 0.289 \\
  \bottomrule
  \end{tabularx}
\end{center}
\end{table}

We compare our method against the results reported in~\cite{wu2023learning}, and against a random pick-and-place policy as a baseline, where pick points are sampled from the object mask and place points randomly within the workspace. Each policy is allowed up to 5 actions, and we report the mean and standard deviation of the normalised improvement over 100 episodes (see Table~\ref{tab:rw_tab}).

Although our method does not surpass~\cite{wu2023learning}, it performs competitively even though it's prediction has not been adjusted to the real-world physics gap via fine-tuning. Direct comparison is limited due to differences in gripper and scene parameters. Nonetheless, it provides a useful reference point for evaluating performance.

\section{CONCLUSION}
We present a novel approach to cloth manipulation that maximally leverages access to the simulation state during training.At an abstract level, our method treats the simulator as a domain in which the agent may conduct \emph{thought experiments} rather than as a faster proxy for the real world. This perspective allows us to optimise our agent's architecture and training to achieve excellent performances with a relatively small network and using simple architectural choices such as convolutional networks and Q-learning. Notably, our results suggest that perception is the dominant bottleneck in cloth manipulation: given full-state supervision, a small model successfully compresses the high-dimensional configuration space into an informative latent vector of size~32. Our findings also highlight the importance of auxiliary objectives in accelerating the training process.

Such a model is not directly transferable to the real world, where full access to the cloth state is unattainable from RGB-D images. Nonetheless, we show that a cross-modality distillation scheme, where the simulation agent teaches a vision-based real-world agent, yields satisfactory zero-shot performance, even if it does not match fine-tuned counterparts. While this may appear limiting, the approach also enables fair comparison of different reasoning modules independently of perception, and likewise comparison of real-world \emph{student} agents irrespective of their simulation teachers.

Our optimisations reduce training time from roughly a week for the MVP baseline to about 40~hours on an NVIDIA RTX~A6000 GPU:~18~hours for pre-training, 6~hours for fine-tuning, and 16~hours for training the real-world agent. We hope that these improvements, together with the modularity of our framework, will accelerate future research in deformable-object manipulation.

Despite these advances, several limitations remain. The pick-and-place action space is restrictive and ill-suited to many cloth-manipulation tasks. Future work will explore finer-grained control for more complex behaviours. Likewise, the square cloth used in our experiments is topologically simple and unrepresentative of most garments.~Graph neural network encoders could capture state representations for objects with more intricate geometry. Finally, our cross-modality transfer relies on hand-crafted resources, limiting scalability. We expect advances in object and scene capture, such as those in~\cite{sam3dteam2025sam3d3dfyimages, delehelle2024garfield, mishra2024closing}, to help automate this step in future.

\addtolength{\textheight}{-6cm}   

\bibliographystyle{IEEEtran}
\bibliography{IEEEabrv,references}

\begin{thebibliography}{10}
\providecommand{\url}[1]{#1}
\csname url@samestyle\endcsname
\providecommand{\newblock}{\relax}
\providecommand{\bibinfo}[2]{#2}
\providecommand{\BIBentrySTDinterwordspacing}{\spaceskip=0pt\relax}
\providecommand{\BIBentryALTinterwordstretchfactor}{4}
\providecommand{\BIBentryALTinterwordspacing}{\spaceskip=\fontdimen2\font plus
\BIBentryALTinterwordstretchfactor\fontdimen3\font minus
  \fontdimen4\font\relax}
\providecommand{\BIBforeignlanguage}[2]{{%
\expandafter\ifx\csname l@#1\endcsname\relax
\typeout{** WARNING: IEEEtran.bst: No hyphenation pattern has been}%
\typeout{** loaded for the language `#1'. Using the pattern for}%
\typeout{** the default language instead.}%
\else
\language=\csname l@#1\endcsname
\fi
#2}}
\providecommand{\BIBdecl}{\relax}
\BIBdecl

\bibitem{matas2018sim}
J.~Matas, S.~James, and A.~J. Davison, ``Sim-to-real reinforcement learning for
  deformable object manipulation,'' in \emph{Conference on Robot
  Learning}.\hskip 1em plus 0.5em minus 0.4em\relax PMLR, 2018, pp. 734--743.

\bibitem{wu2019learning}
Y.~Wu, W.~Yan, T.~Kurutach, L.~Pinto, and P.~Abbeel, ``Learning to manipulate
  deformable objects without demonstrations,'' in \emph{Robotics Science and
  Systems (RSS)}, 2020.

\bibitem{jangir2020dynamic}
R.~Jangir, G.~Alenya, and C.~Torras, ``Dynamic cloth manipulation with deep
  reinforcement learning,'' in \emph{2020 IEEE International Conference on
  Robotics and Automation (ICRA)}.\hskip 1em plus 0.5em minus 0.4em\relax IEEE,
  2020, pp. 4630--4636.

\bibitem{lin2021softgym}
X.~Lin, Y.~Wang, J.~Olkin, and D.~Held, ``Softgym: Benchmarking deep
  reinforcement learning for deformable object manipulation,'' in
  \emph{Conference on Robot Learning}.\hskip 1em plus 0.5em minus 0.4em\relax
  PMLR, 2021, pp. 432--448.

\bibitem{chen2024trakdis}
W.~Chen and N.~Rojas, ``Trakdis: A transformer-based knowledge distillation
  approach for visual reinforcement learning with application to cloth
  manipulation,'' \emph{IEEE Robotics and Automation Letters}, 2024.

\bibitem{salhotra2022learning}
G.~Salhotra, I.-C.~A. Liu, M.~Dominguez-Kuhne, and G.~S. Sukhatme, ``Learning
  deformable object manipulation from expert demonstrations,'' \emph{IEEE
  Robotics and Automation Letters}, vol.~7, no.~4, pp. 8775--8782, 2022.

\bibitem{seita2020deep}
D.~Seita, A.~Ganapathi, R.~Hoque, M.~Hwang, E.~Cen, A.~K. Tanwani,
  A.~Balakrishna, B.~Thananjeyan, J.~Ichnowski, N.~Jamali \emph{et~al.}, ``Deep
  imitation learning of sequential fabric smoothing from an algorithmic
  supervisor,'' in \emph{2020 IEEE/RSJ International Conference on Intelligent
  Robots and Systems (IROS)}.\hskip 1em plus 0.5em minus 0.4em\relax IEEE,
  2020, pp. 9651--9658.

\bibitem{kumar2023offline}
A.~Kumar, R.~Agarwal, X.~Geng, G.~Tucker, and S.~Levine, ``Offline q-learning
  on diverse multi-task data both scales and generalizes,'' in \emph{The
  Eleventh International Conference on Learning Representations}, 2023.

\bibitem{PIGNAT201761}
E.~Pignat and S.~Calinon, ``Learning adaptive dressing assistance from human
  demonstration,'' \emph{Robotics and Autonomous Systems}, vol.~93, pp. 61--75,
  2017.

\bibitem{hoque2020visuospatial}
R.~Hoque, D.~Seita, A.~Balakrishna, A.~Ganapathi, A.~K. Tanwani, N.~Jamali,
  K.~Yamane, S.~Iba, and K.~Goldberg, ``Visuospatial foresight for multi-step,
  multi-task fabric manipulation,'' in \emph{Robotics: Science and Systems},
  2020.

\bibitem{zhang2022learning}
F.~Zhang and Y.~Demiris, ``Learning garment manipulation policies toward
  robot-assisted dressing,'' \emph{Science Robotics}, vol.~7, no.~65, p.
  eabm6010, 2022.

\bibitem{weng2022fabricflownet}
T.~Weng, S.~M. Bajracharya, Y.~Wang, K.~Agrawal, and D.~Held, ``Fabricflownet:
  Bimanual cloth manipulation with a flow-based policy,'' in \emph{Conference
  on Robot Learning}.\hskip 1em plus 0.5em minus 0.4em\relax PMLR, 2022, pp.
  192--202.

\bibitem{lin2022learning}
X.~Lin, Y.~Wang, Z.~Huang, and D.~Held, ``Learning visible connectivity
  dynamics for cloth smoothing,'' in \emph{Conference on Robot Learning}.\hskip
  1em plus 0.5em minus 0.4em\relax PMLR, 2022, pp. 256--266.

\bibitem{yan2021learning}
W.~Yan, A.~Vangipuram, P.~Abbeel, and L.~Pinto, ``Learning predictive
  representations for deformable objects using contrastive estimation,'' in
  \emph{Conference on Robot Learning}.\hskip 1em plus 0.5em minus 0.4em\relax
  PMLR, 2021, pp. 564--574.

\bibitem{ha2022flingbot}
H.~Ha and S.~Song, ``Flingbot: The unreasonable effectiveness of dynamic
  manipulation for cloth unfolding,'' in \emph{Conference on Robot
  Learning}.\hskip 1em plus 0.5em minus 0.4em\relax PMLR, 2022, pp. 24--33.

\bibitem{levine2020offline}
S.~Levine, A.~Kumar, G.~Tucker, and J.~Fu, ``Offline reinforcement learning:
  Tutorial, review, and perspectives on open problems,'' \emph{arXiv preprint
  arXiv:2005.01643}, 2020.

\bibitem{kumar2020conservative}
A.~Kumar, A.~Zhou, G.~Tucker, and S.~Levine, ``Conservative q-learning for
  offline reinforcement learning,'' \emph{Advances in Neural Information
  Processing Systems}, vol.~33, pp. 1179--1191, 2020.

\bibitem{nair2021awacacceleratingonlinereinforcement}
A.~Nair, A.~Gupta, M.~Dalal, and S.~Levine, ``Awac: Accelerating online
  reinforcement learning with offline datasets,'' 2021.

\bibitem{kalashnikov2018scalable}
D.~Kalashnikov, A.~Irpan, P.~Pastor, J.~Ibarz, A.~Herzog, E.~Jang, D.~Quillen,
  E.~Holly, M.~Kalakrishnan, V.~Vanhoucke \emph{et~al.}, ``Scalable deep
  reinforcement learning for vision-based robotic manipulation,'' in
  \emph{Conference on robot learning}.\hskip 1em plus 0.5em minus 0.4em\relax
  PMLR, 2018, pp. 651--673.

\bibitem{ball2023efficientonlinereinforcementlearning}
P.~J. Ball, L.~Smith, I.~Kostrikov, and S.~Levine, ``Efficient online
  reinforcement learning with offline data,'' 2023.

\bibitem{xie2022policyfinetuningbridgingsampleefficient}
T.~Xie, N.~Jiang, H.~Wang, C.~Xiong, and Y.~Bai, ``Policy finetuning: Bridging
  sample-efficient offline and online reinforcement learning,'' 2022.

\bibitem{yang2023foundationmodelsdecisionmaking}
S.~Yang, O.~Nachum, Y.~Du, J.~Wei, P.~Abbeel, and D.~Schuurmans, ``Foundation
  models for decision making: Problems, methods, and opportunities,'' 2023.

\bibitem{van2016deep}
H.~Van~Hasselt, A.~Guez, and D.~Silver, ``Deep reinforcement learning with
  double q-learning,'' in \emph{Proceedings of the AAAI conference on
  artificial intelligence}, vol.~30, 2016.

\bibitem{hasselt2010double}
H.~Hasselt, ``Double q-learning,'' \emph{Advances in neural information
  processing systems}, vol.~23, 2010.

\bibitem{zeng2020transporter}
A.~Zeng, P.~Florence, J.~Tompson, S.~Welker, J.~Chien, M.~Attarian,
  T.~Armstrong, I.~Krasin, D.~Duong, V.~Sindhwani, and J.~Lee, ``Transporter
  networks: Rearranging the visual world for robotic manipulation,''
  \emph{Conference on Robot Learning (CoRL)}, 2020.

\bibitem{clegg2020learning}
A.~Clegg, Z.~Erickson, P.~Grady, G.~Turk, C.~C. Kemp, and C.~K. Liu, ``Learning
  to collaborate from simulation for robot-assisted dressing,'' \emph{IEEE
  Robotics and Automation Letters}, vol.~5, no.~2, pp. 2746--2753, 2020.

\bibitem{loshchilov2017decoupled}
I.~Loshchilov and F.~Hutter, ``Decoupled weight decay regularization,''
  \emph{arXiv preprint arXiv:1711.05101}, 2017.

\bibitem{wu2023learning}
R.~Wu, C.~Ning, and H.~Dong, ``Learning foresightful dense visual affordance
  for deformable object manipulation,'' in \emph{Proceedings of the IEEE/CVF
  International Conference on Computer Vision}, 2023, pp. 10\,947--10\,956.

\bibitem{kingma2014adam}
D.~P. Kingma and J.~Ba, ``Adam: A method for stochastic optimization,''
  \emph{arXiv preprint arXiv:1412.6980}, 2014.

\bibitem{agarwal2021deep}
R.~Agarwal, M.~Schwarzer, P.~S. Castro, A.~Courville, and M.~G. Bellemare,
  ``Deep reinforcement learning at the edge of the statistical precipice,''
  \emph{Advances in Neural Information Processing Systems}, 2021.

\bibitem{sam3dteam2025sam3d3dfyimages}
\BIBentryALTinterwordspacing
S.~D. Team, X.~Chen, F.-J. Chu, P.~Gleize, K.~J. Liang, A.~Sax, H.~Tang,
  W.~Wang, M.~Guo, T.~Hardin, X.~Li, A.~Lin, J.~Liu, Z.~Ma, A.~Sagar, B.~Song,
  X.~Wang, J.~Yang, B.~Zhang, P.~Dollár, G.~Gkioxari, M.~Feiszli, and
  J.~Malik, ``Sam 3d: 3dfy anything in images,'' 2025. [Online]. Available:
  \url{https://arxiv.org/abs/2511.16624}
\BIBentrySTDinterwordspacing

\bibitem{delehelle2024garfield}
D.~Delehelle, D.~Caldwell, and F.~Chen, ``Garfield: Addressing the visual
  sim-to-real gap in garment manipulation with mesh-attached radiance fields,''
  in \emph{2024 IEEE International Conference on Robotics and Biomimetics
  (ROBIO)}.\hskip 1em plus 0.5em minus 0.4em\relax IEEE, 2024, pp. 77--84.

\bibitem{mishra2024closing}
N.~Mishra, M.~Sieb, P.~Abbeel, and X.~Chen, ``Closing the visual sim-to-real
  gap with object-composable nerfs,'' in \emph{2024 IEEE International
  Conference on Robotics and Automation (ICRA)}.\hskip 1em plus 0.5em minus
  0.4em\relax IEEE, 2024, pp. 11\,202--11\,208.

\end{thebibliography}

\end{document}